\documentclass[11pt,a4paper]{article}
\usepackage[hyperref]{ranlp2023}
\usepackage{times}
\usepackage{latexsym}

\aclfinalcopy 

\usepackage{graphicx}
\usepackage{booktabs}
\usepackage{array}
\usepackage{arydshln}
\usepackage{multirow}
\usepackage{microtype}

\usepackage{inconsolata}
\usepackage{xfrac}

\definecolor{mdgreen}{rgb}{0.05,0.6,0.05}

\usepackage[normalem]{ulem}
\def\oddel#1{\bgroup\markoverwith{\textcolor{teal}{\rule[0.4ex]{2pt}{3pt}}}\ULon{#1}}
\def\opdel#1{\bgroup\markoverwith{\textcolor{red}{\rule[0.4ex]{2pt}{3pt}}}\ULon{#1}}
\def\mldel#1{\bgroup\markoverwith{\textcolor{violet}{\rule[0.4ex]{2pt}{3pt}}}\ULon{#1}}

\usepackage{microtype}

\title{With a Little Help from the Authors:\\ Reproducing Human Evaluation of an MT Error Detector}

\author{Ondřej Plátek, Mateusz Lango \and Ondřej Dušek \\
Charles University, Faculty of Mathematics and Physics \\ Institute of Formal and Applied Linguistics \\ Prague, Czech Republic \\
\texttt{\{oplatek,lango,odusek\}@ufal.mff.cuni.cz} \\
}

\date{}

\begin{document}
\maketitle
\begin{abstract}
This work presents our efforts to reproduce the results of the human evaluation experiment presented in the paper of \citet{vamvas-sennrich-2022-little}, which evaluated an automatic system detecting over- and undertranslations (translations containing more or less information than the original) in machine translation (MT) outputs.
Despite the high quality of the documentation and code provided by the authors, we discuss some problems we found in reproducing the exact experimental setup and offer recommendations for improving reproducibility.
Our replicated results generally confirm the conclusions of the original study, but in some cases statistically significant differences were observed, suggesting a high variability of human annotation.
\end{abstract}

\section{Introduction}
\label{sec:intro}
Reproducibility of experimental results is a fundamental principle of scientific research that ensures the validity, credibility, and reliability of scientific findings.
The NLP research community is increasingly interested in reproducibility, which leads to the organization of shared tasks~\cite{belz-etal-2021-reprogen,belz-etal-2022-2022,branco-etal-2020-shared}, the formulation of reproducibility guidelines~\cite{pineau2021improving}, and so on.
However, most previous efforts are limited to the reproducibility of automatic measures, and reproducibility of human evaluation has received less attention~\cite{belz2023missing}.
The ReproHum project,\footnote{\url{https://reprohum.github.io/}} which this paper is a part of, aims to improve this situation.

In this paper, we describe our efforts to reproduce the results of the human evaluation experiment conducted in~\cite{vamvas-sennrich-2022-little} to evaluate the performance of their over- and undertranslation detection method for machine translation (MT).
More specifically, the method detects phrases in the source texts whose meaning is not reflected in an MT output, or phrases in the MT output that are not supported by the source (see details in Sec.~\ref{sec:description}).
The human annotators evaluated the detection accuracy and provided additional reasons for their evaluation by choosing from a list (see Figure~\ref{fig:ui}).
The original experiment was run for English-Chinese and English-German translation pairs, but our reproducibility study is limited to the English-German pair due to the availability of skilled human annotators.

Despite the precise description of the experiment in the original paper, we still encountered difficulties in running the study (see Section~\ref{sec:implementation}), and were only able to finish it successfully thanks to the strong support of the original authors. 
Our results overall support the main claims of \citet{vamvas-sennrich-2022-little}'s original paper, but we still found some discrepancies, despite using the same data, interface and guidelines for the annotation (see Section~\ref{sec:results}).

Our collected annotation outputs, reproduction code, and the filled HEDS sheet \cite{shimorina-belz-2022-human} for the reproduction study are available on Github.\footnote{\url{https://github.com/oplatek/reprohum-as-little-as-possible}}

\section{Original Experiment}
\label{sec:description}

The original paper~\cite{vamvas-sennrich-2022-little} proposed an automatic method for detecting coverage errors in the output of MT systems.
Coverage errors include \emph{undertranslations}, i.e. the omission of important source content in the MT system output, and \emph{overtranslations}, i.e. the addition of superfluous words to the translation that may not be supported by the source.

The method uses contrastive conditioning~\cite{contrastive} and finds coverage errors by iteratively computing the probability of the generated translation with an MT system conditioned on an incomplete text source.
If the probability of the generated translation increases when a particular phrase is deleted from the source, the method takes this as an indicator that the deleted phrase is not adequately reflected in the translation and treats it as an omission (undertranslation).
Similarly, by reversing the source with the target, the method also detects overtranslations.
To summarize, for a given input-translation pair, the output of the method is the type of problem detected (over- or undertranslation) and a phrase that has been omitted from the translation (in the case of undertranslation), or that is superfluous (in the case of overtranslation).

\begin{figure*}[ht]
    \centering
    \includegraphics[width=\linewidth]{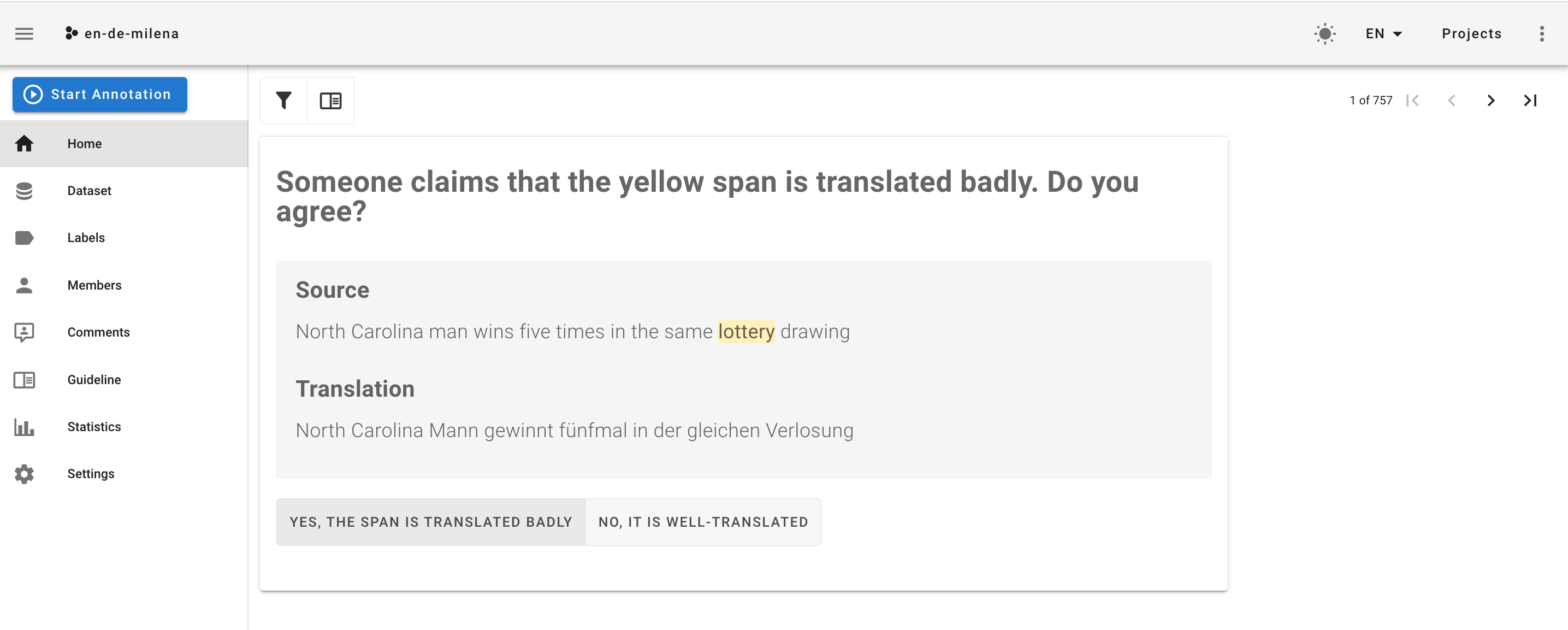}

    \noindent
    \includegraphics[width=0.49\linewidth]{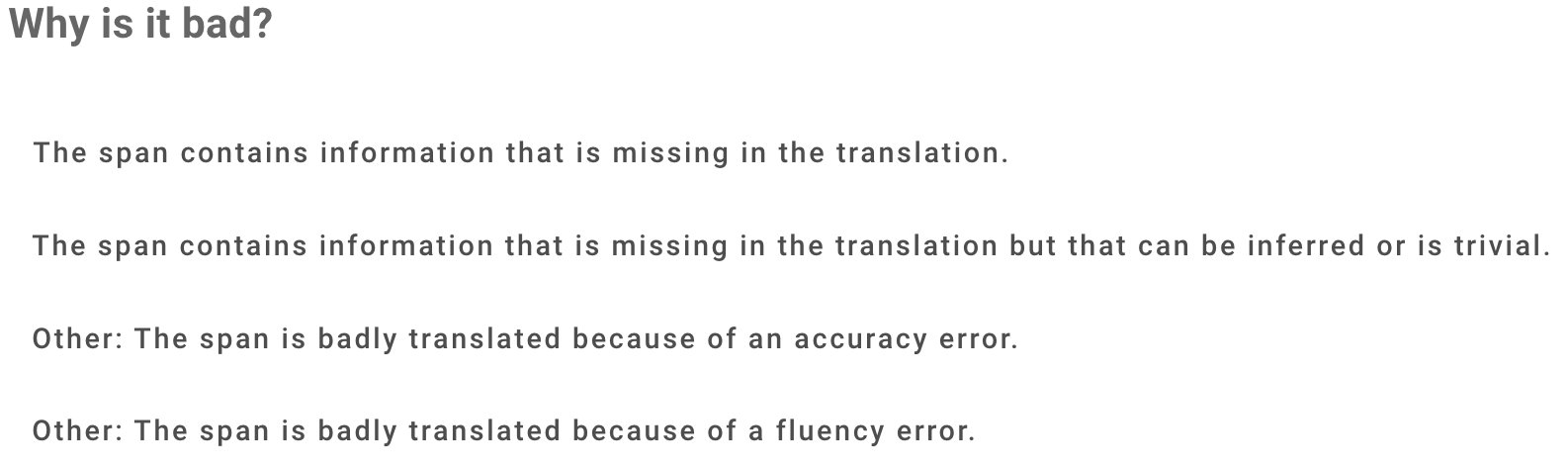}
    \includegraphics[width=0.49\linewidth]{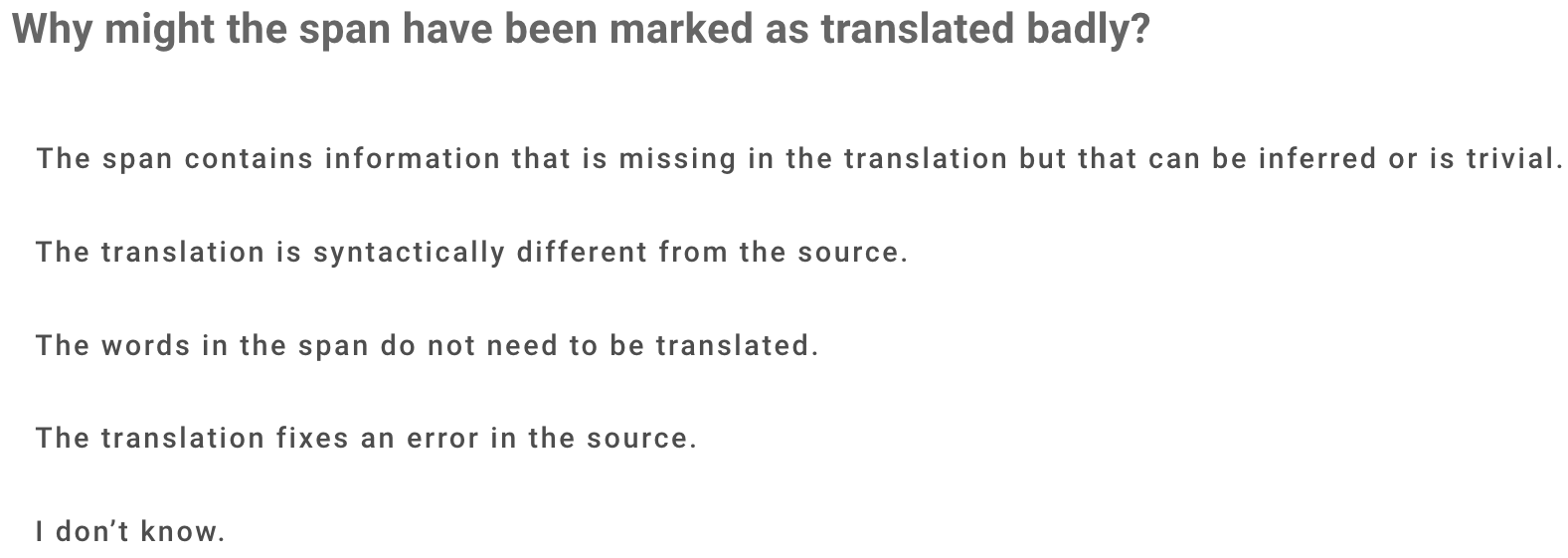}
    \caption{Screenshots of the Annotation Interface where the annotator needs to: (1) Top image; Select whether the source sentence is well translated. (2) Bottom left image; In the case of a bad translation, indicate the type of error. (3) Bottom right image; In the case of a correct translation, hypothesize why it was marked as an error.}
    \label{fig:ui}
\end{figure*}

The corresponding human evaluation aimed to analyze in detail the predicted problematic text spans and assess their correctness.
Human annotators were presented with a source sentence, the generated MT translation, and a highlighted passage.
The annotator's task was to decide whether the highlighted passage was correctly translated and later to select additional feedback for fine-grained analysis from a given single-choice list.
If the annotator confirmed that the highlighted span was incorrectly translated, they were asked to specify the type of error (e.g., fluency error, accuracy error, addition/omission of non-trivial information, etc).
On the other hand, if the annotator considered the span to be correctly translated, they were asked to give a possible reason why one could think that it was translated incorrectly (e.g., syntactic differences, adding/removing trivial/obvious information).
The full list of possible reasons can be seen in Figure~\ref{fig:ui}.

The manual evaluation was carried out by two linguists who were provided with a two-page document containing annotation guidelines.
The guidelines included the task description, instructions on using the annotation interface, and examples of three incorrect and three correct translations.
Each annotator responded to approximately 700 randomly selected examples.

\section{Differences in Our Reproduction Study}

We aimed to conduct the reproduction as close as possible to the original study. We worked on the same set of system outputs, with the identical annotation interface and instructions.\footnote{\url{https://github.com/ZurichNLP/coverage-contrastive-conditioning}}
However, there were some differences with respect to annotator hiring and to splitting the annotation between them.

\paragraph{Annotator hiring}
We hired two annotators who were university students and native speakers of German with high proficiency in English, same as in the original study.
We used contacts arranged through ReproHum organizers at two German universities (Bielefeld and Munich), which means that the students spoke a different variety of German from the original study, which was conducted in Zurich, i.e., with Swiss German speakers.
In addition, one of our annotators was from a different study field (public health) than the original study's annotators (NLP).
The reason is that we could not find interested NLP students at the time of hiring.
Each annotator was paid €180 for the approximate 10 hours of work. The €18 hourly wage differs from the original study (which reports ca.~\$30) but is in line with ReproHum recommendations (150\% of German minimum wage).

\paragraph{Data split for annotation}
We used the same input data for the human evaluation, i.e., the same outputs of the machine translation (MT) system, together with error annotations.
Similarly to the original study, the sentences with annotated errors were split randomly into two parts, for one annotator each.
However, the data in each split were presented to each of the annotators in a different random order.

\section{Implementation Issues}
\label{sec:implementation}


We found two implementation problems while running the study: one with setting up the annotation interface and the other with the script computing final statistics.

\paragraph{Annotation interface}


The authors of the original study used the popular open-source annotation software Doccano~\cite{doccano}, which was customized to implement the interface required for their human evaluation experiment.
The original open-source software has been updated over time, making the authors' customization incompatible with the toolkit.
Even after downgrading Doccano to the version used by the authors, some of the dependencies were found to be no longer available.
Our attempts to use newer versions of dependencies and/or Doccano were unsuccessful.

Finally, with the permission of the ReproHum organisers, we contacted the authors, who fortunately kept the annotation interface in an easy-to-distribute form of a Docker snapshot~\cite{merkel2014docker}.
With the Docker image provided, we were able to run the reproduction with the identical annotation interface.

\paragraph{Statistics computation}
To compute the necessary annotation statistics, we used the evaluation script provided by \citet{vamvas-sennrich-2022-little}.
During our data analysis, we noticed that the script did not correctly handle examples with multiple spans annotated within the same sentence.
For such examples, the last annotation analysed by the script would override the previous ones, resulting in some annotations being unintentionally removed.
We fixed this bug and ran the analysis with the original and corrected script on the annotation data from both the original and repeated study (see Table~\ref{tab:statistics}).

\section{Results}
\label{sec:results}

A side-by-side presentation of all the results can be found in Table~\ref{tab:statistics}.
We report results calculated using both the original script (Original, Reproduced) and the corrected script detailed in Section~\ref{sec:implementation} (Original\_v2, Reproduced\_v2).
The use of the corrected script increased the number of examples by 54 for the original results and by 56 for the reproduced results, which represents less than 4\% of the total data size.
In order to maintain a better reference to the results reported in the original paper, and since the difference is not large, the following discussion will analyse the results computed with the original script.

In both experiments, each annotator provided feedback on approximately 700 examples, which were similarly divided into examples of over- and under-translation.
The inter-annotator agreement, measured with Cohen's $\kappa$, was higher in our reproduction for the simple evaluation of the correctness of highlighted spans, but the results were within the constructed 95\% confidence interval for the original value.
According to~\citet{mchugh2012interrater}, these $\kappa$ values should be interpreted as weak/moderate agreement.
On the other hand, the inter-annotator agreement for fine-grained responses was statistically significantly lower in our repeated study, but again, both values from the original and the repeated study should be interpreted as “minimal agreement” \cite{mchugh2012interrater}.

\begin{table*}[htbp]
\centering\small
\begin{tabular}{p{8cm}cccc}
\toprule
 & \bf Original & \bf Reproduced & \bf Original\_v2 & \bf Reproduced\_v2 \\
\multicolumn{5}{c}{\bf Basic statistics}\\
\midrule
Annotator 1 no. of overtranslation examples & 372 & 372 & 390 & 389 \\
Annotator 1 no. of undertranslation examples & 344 & 348 & 354 & 361 \\
Annotator 1 total no. of examples& 716 & 720 & 744 & 750 \\
\midrule
Annotator 2 no. of overtranslation examples & 348 & 352 & 362 & 366 \\
Annotator 2 no. of undertranslation examples & 351 & 350 & 363 & 362 \\
Annotator 2 total no. of examples & 699 & 702 & 725 & 728 \\
\midrule
\multicolumn{5}{c}{\bf Inter-annotator agreement}\\
\midrule
Number of overlapping samples & 466 & 471 & 474 & 479 \\
Cohen's $\kappa$ for trans. correctness eval.  & 0.5343 & 0.6044 & 0.5593 & 0.6313 \\
Cohen's $\kappa$ for fine-grained answers & 0.3186 & 0.2420 & 0.3328 & 0.2536 \\
\midrule
\multicolumn{5}{c}{\bf Precision of spans indicated by the method as incorrect }\\
\midrule
Incorrectly translated spans detected as overtanslations& 49 & 45 & 54 & 48 \\
Correctly translated spans detected as overtanslations & 611 & 619 & 638 & 647 \\
Incorrectly translated spans detected as undertranslations & 240 & 135 & 250 & 143 \\
Correctly translated spans detected as undertranslations & 369 & 476 & 379 & 491 \\
\midrule
\multicolumn{5}{c}{\bf Fine-grained analysis for overtranslations}\\
\midrule
True overtranslations - addition of trivial or inferable information & 10 & 2 & 10 & 2 \\
True overtranslations - addition of unsupported information & 5 & 11 & 5 & 12 \\
True errors - accuracy errors & 28 & 24 & 32 & 24 \\
True errors - fluency errors & 6 & 8 & 7 & 10 \\
\midrule
False errors - addition of redundant but fluent info. & 113 & 120 & 117 & 124 \\
False errors - addition of supported information & 19 & 30 & 20 & 30 \\
False errors - a syntactic difference & 428 & 254 & 449 & 263 \\
False errors - unknown reason & 51 & 215 & 52 & 230 \\
\midrule
\multicolumn{5}{c}{\bf Fine-grained analysis for undertranslations}\\
\midrule
True undertranslations - lack of important information & 114 & 80 & 120 & 86 \\
True undertranslations - lack of redundant information & 107 & 7 & 110 & 7 \\
True errors - accuracy errors  & 16 & 35 & 17 & 37 \\
True errors - fluency errors & 3 & 13 & 3 & 13 \\
\midrule
False errors - fluency errors in the source & 72 & 107 & 72 & 110 \\
False errors - addition of redundant but fluent info.& 25 & 103 & 25 & 104 \\
False errors -  a syntactic difference & 249 & 174 & 257 & 178 \\
False errors - unknown reason  & 23 & 92 & 25 & 99 \\
\bottomrule
\end{tabular}
\caption{The summary of the raw results obtained in the original and reproduced study.
The columns with “v2” suffix are computed with the fixed evaluation script (see Sec.~\ref{sec:implementation}).
Note that annotators for Original and Reproduced studies differ: Annotator 1 is a different different person for the Original(\_v2) and Reproduced(\_v2) studies. Similarly for Annotator 2.}
\label{tab:statistics}
\end{table*}

\begin{table*}[tp]
\begin{center}
\small
\begin{tabular}{llrcrr}
\toprule
                        &                 & \bf Original  & \bf 95\% CI            & \bf Reproduction  &  \bf CV*   \\
\midrule
\multirow{2}{*}{\bf Target} & Addition errors & 2.3            & (1.38; 3.71)   & 1.95  & 16.42 \\
                        & Any errors      & 7.4            & (5.66; 9.68)   & 6.77  & 8.86  \\
\midrule
\multirow{2}{*}{\bf Source} & Omission errors & 36.3           & (32.57; 40.18) & * 14.23  & 19.56 \\
                        & Any errors      & 39.4           & (35.61; 43.34) & * 22.09  & 15.34\\
\bottomrule
\end{tabular}
\end{center}
\caption{Word-level precision (\%) of the spans that were highlighted by the method \cite[Table 2]{vamvas-sennrich-2022-little} in the original study and in our reproduction, together with 95\% confidence intervals constructed for the original values (95\% CI)  and the small-sample coefficient of variation (CV*). Reproduced results that do not fall within the CI are marked with an asterisk. }
\label{tab:precision}
\end{table*}

In the original paper, the results of the experiments are reported in terms of word-level precision of the highlighted spans.
The comparison of the original precision values with those obtained in our reproduction can be observed in Table~\ref{tab:precision}.
In addition, for each original value, we computed a 95\% confidence interval using Wilson's score method~\cite{afed75f6-c9dc-3279-8854-99fa262b33b1}.
The precision values obtained in our reproduction are generally lower than those reported in the original study.
However, the differences in precision for the overtranslation spans are still within the confidence intervals.
In contrast, the differences for under-translation are substantial, as the reproduced precision values are about 44-46\% lower.
This difference is also statistically significant at the significance level $\alpha=5\%$.

\begin{table}[t]
\centering\small
\begin{tabular}{llrrr}
\toprule
                                  &                  &   \bf $\chi^2$     & \bf p-value & $V$          \\
\midrule
\multirow{2}{*}{Overtrans.}  & good trans. & 355.77 & \textless{}0.0001& 0.50\\
                                  & bad trans.  & * 201.88 & \textless{}0.0001&0.71 \\
\midrule
\multirow{2}{*}{Undertrans.} & good trans. & 596.99 & \textless{}0.0001 &0.57\\
                                  & bad trans.  & * 15.8   & 0.0016    &    0.34    \\
\bottomrule
\end{tabular}
\caption{The results of goodness-of-fit (GOF) tests of human answers in fine-grained analysis (types of error) in the original and our reproduced study. The effect size is measured with Cramér's $V$ for GOF. For cases marked with an asterisk, the conditions to use $\chi^2$ approximation were not met, thus the test statistics were estimated with Monte Carlo simulation (10k samples).  }
\label{tab:chi2}
\end{table}

To compare the results of the fine-grained analysis, we performed $\chi^2$ goodness-of-fit tests between the answers provided by the original annotators and those provided by us, as well as calculating Cramér's $V$ for goodness-of-fit.
The results are presented in Table~\ref{tab:chi2}.
We were able to reject the null hypothesis that the reproduced fine-grained responses follow the distribution of the original responses with low p-values for all four sets of results.
All obtained values of Cramér's $V$ exceed the 0.29 threshold suggested by~\citet{cohen1988spa} as an indicator of a large discrepancy between the data distributions.

The differences are visible even to the naked eye, as our annotators selected unknown reasons for highlighting a correctly translated text span about four times more often than in the original study.

In the case of undertranslations, the annotators in the original study chose much more often  that the translation is incorrect, but the missing information can be inferred or is trivial (107 vs.\ 7 counts).
On the other hand, our annotators were much more likely to consider that the translation was correct but could be considered inaccurate because some trivial or easily inferable information was missing (25 vs.\ 103 counts).
Our annotators also found more spans highlighted as under-translations as reasons for accuracy or fluency errors in the translation.
These larger differences for examples of undertranslation may also indicate that this variant of the evaluation task is more difficult.
Note that the annotators have a highlighted text span in the source text, but still have to answer questions about the final translation without any word/phrase alignment information.

\section{Quantifying Reproducibility}
\label{sec:quant}

Following the guidelines of the ReproHum shared task \cite[Sect.~A5]{belz2023missing}, we identify reproduction targets in the following categories:
\begin{itemize}
\item Type I -- numerical scores:
\begin{itemize}
\item the precision of text spans labeled as over-/undertranslations to truly contain over-/undertranslation errors
\item the precision of text spans labeled as over-/undertranslations to contain some translation errors
\end{itemize}
\item Type II -- sets of numerical values:
\begin{itemize}
\item the set of precision results for examples marked as overtranslations
\item the set of precision results for examples marked as undertranslations
\end{itemize}
\item Type III -- categorical labels attached to text spans:
\begin{itemize}
\item Sets of spans annotated with the correct/incorrect translation label, separately for over- and under-translations.
\item Sets of fine-grained reasons given by annotators for marking a span as incorrect, separately for over- and under-translations and for correctly/incorrectly detected spans.
\end{itemize}
\end{itemize}

\paragraph{Type I} For the numerical results, we followed the quantified reproducibility assessment by~\citet{belz-etal-2022-quantified}, which 
involves
calculating the small sample coefficient of variation (CV*) as a measure of the degree of reproducibility.
The results are given in the last column of Table~\ref{tab:precision}.
Three out of four CV* values are in the 15-20 range.
Only for the precision of detecting a translation error in the text span marked as overtanslation, the CV* value is significantly lower (8.86).

\paragraph{Type II} results are usually evaluated with Pearson's correlation~\cite{huidrom-etal-2022-two}, but there is little point in calculating it here.
It is equal to 1 for both over- and under-translations, while the standard statistical test for correlation fails to reject the null hypothesis that the true correlation is 0.

\begin{table}[t]
\centering\small
\begin{tabular}{lcc}
\toprule
&$\alpha$&\bf \%Ident.\\\midrule
Overtranslation  & 0.6976 & 0.9558 \\
Undertranslation & 0.3762 & 0.7266 \\\midrule
Joint            & 0.5109 & 0.8475 \\\bottomrule
\end{tabular}
\caption{Krippendorff's alpha coefficient ($\alpha$) for assessment of bad/good translation using combined annotations for both original and replication studies. \%Ident.\ is the percentage of identical answers between the original and replicated annotation. (The annotations of both annotators were combined into one list. For examples where both annotators provided an answer, half of the answers were taken from each annotator.)}
\label{tab:alpha-general}
\end{table}

\begin{table}[t]
\centering\small
\begin{tabular}{llrr}
\toprule
&&$\alpha$&\bf \%Ident.
\\\midrule

\multirow{3}{*}{Overtranslation}  & Good translation & 0.2238&0.5059 \\
                                  & Bad translation  & 0.1982  &0.4687\\
                                  & Joint            & 0.2607 &0.5033\\\midrule
\multirow{3}{*}{Undertranslation} & Good translation & 0.1427 &0.3365 \\
                                  & Bad translation  & 0.1994 & 0.4468\\
                                  & Joint            & 0.2084 & 0.3621\\\midrule
Joint                             &                  & 0.2664 & 0.4366\\\bottomrule
\end{tabular}
\caption{Krippendorff's alpha coefficient ($\alpha$) for fine-grained analysis using the combined annotations (see comment in Table~\ref{tab:alpha-general}).  \%Ident.\ is the percentage of identical answers between the original and replicated annotation.
 }
 \label{tab:alpha-fine}
\end{table}

\paragraph{Type III} Finally, the reproducibility of categorical labels was assessed with Krippendorff’s alpha.
Since the aim of this analysis is not to measure the agreement between all four annotators (2 from the original study and 2 from the replication) but rather to measure the reproducibility, for the purpose of computing Krippendorff’s alpha the annotations obtained from each pair of annotators were combined into one set.
For the overlapping examples i.e. examples annotated by both annotators, the response of a randomly chosen annotator was retained.

The values of Krippendorff’s alpha  together with the percentage of identical evaluations in both the original and reproduced study for coarse-grained annotations (i.e. correct/incorrect translation label) are given in Tab.~\ref{tab:alpha-general}.
The value for overtranslation is significantly higher than for undertranslations and is  above the $\sfrac{2}{3}$ threshold suggested by~\citet{krippendorff2004content} as the lowest limit to consider a good agreement between the raters.
The high agreement between the original annotation and the reproduced one can also be observed in terms of the proportion of identical answers -- almost 85\% for the whole dataset.

Similar results for fine-grained annotations are provided in Tab.~\ref{tab:alpha-fine}.
All reported values of Krippendorff’s alpha are relatively low, and the proportion of examples that are evaluated identically in both studies is below 51\%.
However, the ratio for overtranslation spans marked as correct translations is significantly higher than the same ratio for undertranslations (roughly 16 percentage points).

\section{Findings}
\label{sec:findings}
Based on the manual evaluation, the authors of the original paper present several findings/conclusions:
\begin{itemize}
\item Precision is higher for undertranslations, but still low for overtranslations
\item Many of the highlighted spans are translation errors, but not over/undertranslations
\item Fine-grained analysis suggests that syntactic differences contribute to the false positives for overtranslations.
\end{itemize}

All of the above findings roughly correspond to the results of our reproduced experiment.
The precision for undertranslations is also higher than for overtranslations, but the difference between the two was considerably smaller in our experiment.
For example, the difference in precision for true coverage errors is 34.0 percentage points in the original study but only 12.3 in ours.
Similarly, our fine-grained analysis confirms that syntactic differences contribute to false positives, but they are reported about 40\% less frequently than in the original study.
However, as mentioned above, this is partly due to the fact that our annotators were much more likely to select the “unknown reason” response.


\section{Conclusion}

We carefully repeated a human evaluation study of paper~\cite{vamvas-sennrich-2022-little}.
Despite the high-quality documentation and well-organized code provided by the authors, we encountered several problems that were difficult to overcome.
In particular, we would not have been able to run the annotation interface and repeat the study without the authors' help.
We also noticed a minor error in their evaluation script and suggested a modification.
The reproduction process made us realise that it is almost impossible to publish a fully reproducible paper without actually trying to reproduce it end-to-end.
We also advocate distributing annotation interfaces in the form of a Docker image containing all the dependencies.

Our overall results agree with the high-level conclusions made in the original paper.
The reproduction of results of the coarse-grained analysis was successful for overtranslations, but the results for undertranslations were significantly lower than the reported in the original study.
The results of the fine-grained analysis were even more difficult to reproduce -- we observed significant differences in all analyzed groups of answers.
This may suggest that when designing experiments with human judgments, setups with a very limited number of possible answers (especially binary questions) are easier to replicate and should be prioritized over more complex setups whenever possible.


\section*{Acknowledgments}
This research was supported by Charles University projects GAUK 40222 and SVV 260575 and by the European Research Council (Grant agreement No.~101039303 NG-NLG).
It used resources provided by the LINDAT/CLARIAH-CZ Research Infrastructure (Czech Ministry of Education, Youth, and Sports project No. LM2018101).
The authors are very grateful to Jannis Vamvas and Rico Sennrich for assisting in this reproduction study.
We also thank Steffen Eger and Ivan Habernal for helping us find annotators.

\bibliographystyle{acl_natbib}
\bibliography{ranlp2023}

\end{document}